\documentclass[10pt,twocolumn,letterpaper]{article}

\usepackage{wacv}
\usepackage{times}
\usepackage{epsfig}
\usepackage{graphicx}
\usepackage{amsmath}
\usepackage{amssymb}
\usepackage{booktabs}
\usepackage{cuted}
\usepackage{capt-of}
\usepackage{makecell}
\usepackage[accsupp]{axessibility}

\usepackage{cite}
\usepackage{pifont}
\newcommand{\cmark}{\ding{51}}
\newcommand{\xmark}{\ding{55}}

\usepackage{caption,setspace}
\usepackage{subcaption}
\def\wacvPaperID{1358} % Enter the WACV Paper ID here

\wacvalgorithmstrack   % Uncomment this line if you are submitting to the Algorithms Track.
\wacvfinalcopy % *** Uncomment this line for the final submission

\ifwacvfinal
\usepackage[breaklinks=true,bookmarks=false]{hyperref}
\else
\usepackage[pagebackref=true,breaklinks=true,colorlinks,bookmarks=false]{hyperref}
\fi

\begin{document}

%%%%%%%%% TITLE
\title{Watching the News: Towards VideoQA Models that can Read}
 
% Towards VideoQA Models That Can Read And Watch News Videos
% Did you pay attention? Towards VideoQA Models that can Read
% Did you understand the news? Towards VideoQA Models that can Read
% Are you sure you understand? Towards VideoQA Models that can Read
% Read as you watch: Towards VideoQA Models that can Read
% \task: Towards VideoQA Models that can Read

\author{\quad Soumya Jahagirdar$^{\dagger}$  \qquad Minesh Mathew$^{\dagger}$ \qquad Dimosthenis Karatzas$^{\ddagger}$ \qquad C. V. Jawahar$^{\dagger}$ \\
{\tt\footnotesize \{soumya.jahagirdar, minesh.mathew\}@research.iiit.ac.in} \enspace \enspace {\tt\footnotesize dimos@cvc.uab.es}  \enspace\enspace {\tt\footnotesize jawahar@iiit.ac.in} \\
$^{\dagger}$ CVIT, IIIT Hyderabad, India \enspace\enspace
         $^{\ddagger}$ Computer Vision Center, UAB, Spain}

% \author{Soumya Jahagirdar{1}\\
% {\tt\small soumya.jahagirdar@research.iiit.ac.in}
% \and
% Minesh Mathew\\
% {\tt\small minesh.mathew@research.iiit.ac.in}

% \and
% Dimosthenis Karatzas\\

% \and
% C.V. Jawahar\\

\maketitle
\thispagestyle{empty}

% \begin{strip}\centering
% \includegraphics[width=\textwidth]{images/task.png}
% \captionof{figure}{NewsVideoQA dataset: A new dataset Towards VideoQA Models that can Read.
% \label{fig:task}}
% \end{strip}

%%%%%%%%% ABSTRACT
\begin{abstract}
  Video Question Answering methods focus on commonsense reasoning and visual cognition of objects or persons and their interactions over time. Current VideoQA approaches ignore the textual information present in the video. Instead, we argue that textual information is complementary to the action and provides essential contextualisation cues to the reasoning process. To this end, we propose a novel VideoQA task that requires reading and understanding the text in the video. To explore this direction, we focus on news videos and require QA systems to comprehend and answer questions about the topics presented by combining visual and textual cues in the video. We introduce the ``NewsVideoQA'' dataset that comprises more than $8,600$ QA pairs on $3,000+$ news videos obtained from diverse news channels from around the world. We demonstrate the limitations of current Scene Text VQA and VideoQA methods and propose ways to incorporate scene text information into VideoQA methods. 
\end{abstract}

% \vspace{-0.5cm}

%In this work, we propose a new VideoQA task that requires reading and understanding textual information in the video. Textual information is complementary to the action, providing essential contextualisation cues to the reasoning process. 
 %Furthermore, we thoroughly analyze baseline and demonstrate the methods by adapting existing Scene Text VQA and Video QA models to this new domain. 
 
%%%%%%%%% BODY TEXT
\section{Introduction}

%DK TRY
%TASK:

Visual Question Answering has evolved in numerous directions over the past few years. Two promising directions are, on one hand, the attempt to apply VQA on more dynamic scenarios, namely on video inputs and on the other hand, the introduction of scene text as an extra modality in the VQA process. 

\begin{figure} 
    \centering
    \includegraphics[width=1\linewidth]{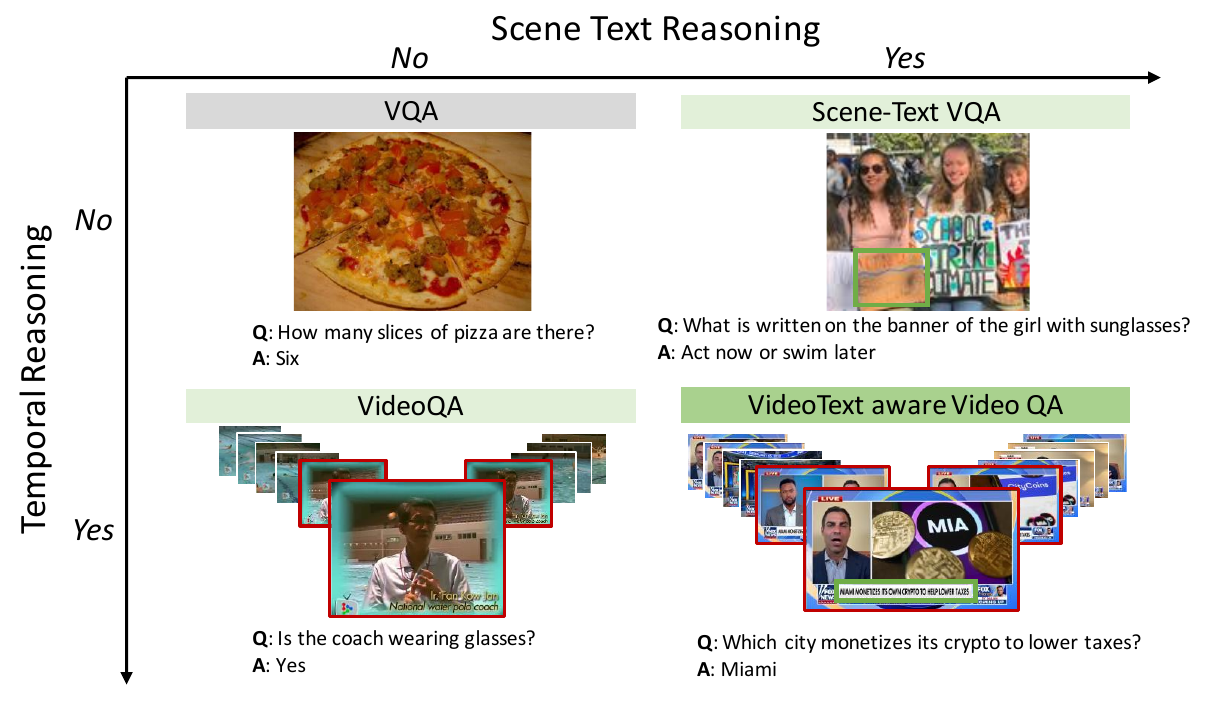}
    \caption{\small We address the task of text based Video Question Answering, incorporating VideoText (VideoText is the textual content embedded in the videos) information (bottom right). We propose a new dataset of News Videos along with QA annotations grounded on video text, and explore VQA models that jointly reason over temporal and text based information.}
    \label{fig:task}
\end{figure}

The reasoning processes required to tackle these challenges are not trivial to incorporate into a model. Taking into account the temporal dimension of an unfolding event requires reasoning over the evolution of certain actions, retrieving information from a specific time in the sequence, or a combination of the two. At the same time, recognizing the fact that world around us is littered with textual information that often carries important semantics necessary to interpret the scene has spawned a new direction in VQA. Introducing the scene text modality in the process requires incorporating error-prone reading systems, and connecting scene text semantics and literal transcriptions with the answer space.

In this work, we attempt for the first time to join these two lines of research, and introduce the VideoText (VideoText is the textual content embedded in the videos) modality into Video Visual Question Answering.

%STATE OF THE ART:
%FLAW IN SOA:

Various attempts to apply VQA to the video setting have been proposed~\cite{newskvqa,  tvqa, moviesqa, refined_attention, activityqa}. Such VideoQA methods have put forward datasets and methods focusing on recognizing actions, emotions, activities, and reasoning over temporal, causal correspondences, and knowledge graphs. However, they fall short in reasoning over the text appearing in the videos. 

Scene Text VQA \cite{textvqa, stvqa}, on the other hand, focuses on methods that allow VQA systems to incorporate scene text in the reasoning process. On one hand, this entails extracting semantics from noisy textual input, and on the other hand it requires dynamically expanding the answer space to incorporate new answer tokens afforded by the scene text \cite{textvqa, stvqa, doccvqa, docvqa, ocrvqa}. Nevertheless, all scene text VQA methods are limited to processing a single image and cannot be readily extended to a multi-frame video input.

%YOUR IDEA / SOLUTION:

In this work, we attempt to combine multi-frame based, VideoQA architectures with the scene text modality (Fig.~\ref{fig:task}). To explore this novel research direction, we define a new task and associated dataset: NewsVideoQA. Motivated by the prominent function of scene text in news video snippets, and the complementary information it carries to the visual modality, we consider that Visual Question Answering over News Videos is an adequate task to advance in models that jointly reason over temporal and scene text based information.

%\begin{figure}
%    \centering
%    \includegraphics[width=1\linewidth]{images/task.png}
%    \caption{\small Existing datasets for video question answering do not consider text in the video to answer questions. Given that textual information is an important component for improved video comprehension, we propose a new task of text-based video question answering followed by a new dataset, \textbf{NewsVideoQA}, for the proposed task. }
%    \label{fig:task}
%\end{figure}

%PROOF IT WORKS:
We present and thoroughly analyse the NewsVideoQA dataset, indicating key statistics and theoretical upper bound performance in various scenarios. We subsequently explore various baseline methods and demonstrate the limitations of both VideoQA and Scene Text VQA methods. We show that Scene Text VQA methods only yield top performance when they are applied on the video frame corresponding to the question (that includes the information needed to answer), but there is no trivial way for such methods to automatically retrieve the right frame. On the other hand, we show that VideoQA methods that do not consider the scene text, result in very low performance on the NewsVideoQA dataset. Finally, we repurpose a recently proposed VideoQA method to incorporate scene text information and show that it yields top results on the NewsVideoQA dataset, combining the benefits of both VideoQA and Scene Text VQA genres. The dataset is available at \href{http://cvit.iiit.ac.in/research/projects/cvit-projects/videoqa}{http://cvit.iiit.ac.in/research/projects/cvit-projects/videoqa}

The contributions of our work are the following:
\begin{itemize}
    \item We introduce a new task of text based Video Question Answering, in which models must have the ability to read and reason about textual content in the videos (multi-frame input) to answer questions.
    \item We propose a new dataset: NewsVideoQA to explore the proposed task. This dataset comprises questions defined over the textual content in news videos and requires models to read and reason over it to obtain an answer.
    \item We evaluate various baselines on the NewsVideoQA dataset. These baselines include simple heuristic methods, text-only (machine comprehension) models, Scene Text VQA and VideoQA models.
	\item We repurpose the SINGULARITY~\cite{singularity} VideoQA model to the NewsVideoQA task and yield acceptable results compared to the original model.
\end{itemize}

\begin{table*}[]
    \centering
     % \footnotesize
    \small
    \caption{\textbf{A comparative overview of VideoQA datasets.} Datasets prior to our work, consider video, video + subtitles, video + knowledge base as input. Our work introduces a new line of research where the questions in proposed dataset are framed based on textual content in the news videos. The column Synthetic Gen. indicates the dataset which are synthetically/automatically generated.} 
    \begin{tabular}{@{}lccrccrr@{}}\toprule
        \textbf{Dataset}  & \textbf{Subtitles} & \textbf{Text in video} & \textbf{Type of videos} & \textbf{Synthetic Gen.} & \textbf{Free-form} & \textbf{\#Video} & \textbf{\#QA} \\ 
        \midrule
        VideoQA \cite{uncovering_temporal_vidqa} & \textcolor{red}{\xmark} & \textcolor{red}{\xmark} & Cooking, movies & \textcolor{green}{\cmark} &  \textcolor{red}{\xmark} & 109K & 390K   \\
        
        MSVD-QA \cite{refined_attention} & \textcolor{red}{\xmark} & \textcolor{red}{\xmark} & YouTube & \textcolor{green}{\cmark} & \textcolor{green}{\cmark} & 1.9K & 50K  \\
        
        ActivityNet-QA \cite{activityqa} & \textcolor{red}{\xmark} & \textcolor{red}{\xmark} & YouTube & \textcolor{red}{\xmark} & \textcolor{green}{\cmark}  & 5.8K & 58K \\
        
        MSRVTT-QA \cite{refined_attention} &  \textcolor{red}{\xmark} & \textcolor{red}{\xmark} & YouTube & \textcolor{green}{\cmark} & \textcolor{green}{\cmark} & 10K & 243K \\
        
        MoviesQA \cite{moviesqa} &  \textcolor{green}{\cmark} & \textcolor{red}{\xmark} & Movies & \textcolor{red}{\xmark} & \textcolor{red}{\xmark} & 6.7K  & 6.4K \\
        
        TVQA \cite{tvqa} & \textcolor{green}{\cmark} & \textcolor{red}{\xmark} & TV shows & \textcolor{red}{\xmark} & \textcolor{red}{\xmark} & 21K & 152K  \\
        
        HowtoVQA69M \cite{just_ask} & \textcolor{green}{\cmark} & \textcolor{red}{\xmark} & TV shows & \textcolor{green}{\cmark} & \textcolor{red}{\xmark} & 69M & 69M  \\
        
        QA News Videos \cite{old_news_videoqa} &  \textcolor{red}{\xmark} & \textcolor{green}{\cmark} & Web videos & - & - & - & 40 \\ 
        
        NewsKVQA \cite{newskvqa}  & \textcolor{green}{\cmark} & \textcolor{red}{\xmark} & News videos & \textcolor{green}{\cmark} & \textcolor{red}{\xmark} & 5.8K & 58K \\ 
        
        \textbf{NewsVideoQA (Ours)}  & \textcolor{green}{\cmark} & \textcolor{green}{\cmark} & {News videos} & \textcolor{red}{\xmark} & \textcolor{green}{\cmark} & {\begin{tabular}[c]{@{}l@{}}3.0K \\ \end{tabular}} & {8.6K} \\ 
        \bottomrule
    \end{tabular}
    \label{tab:dataset_table}
\end{table*}

\section{Related Work}
\label{sec:related_works}
% With the growing interest in vision and language problems, there have been many works that deal with question answering on images and videos. 
In this section, we briefly discuss some essential works in this space that is relevant to our work.

\textbf{Video Question Answering. } 
%With the growing importance of integrating vision and language, many works have tackled the problem of obtaining the answer given an video and a question.
% With the growing interest in vision and language problems, there have been many works that deals with the problem of Video QA--- answering questions asked on videos.
One of the early attempts at VideoQA is a retrieval-based approach for factoid QA proposed by Yang et al.~\cite{old_news_videoqa}. Their system relies on speech transcripts and external knowledge to answer the questions. One or more sentences from the transcript are returned as the output of the QA system, and the output is considered correct if the target answer is contained within the  retrieved sentences. For QA evaluation, they used a private dataset containing only 40 QA pairs. Contrary to this work, our NewsVideoQA focuses particularly on the text appearing in the news videos, and is defined over a much larger dataset.

% Similar to their work, QA in this work deals with news videos. However, we  introduce a new video QA task and a much larger dataset where all questions are based on the  text seen in the videos. Unlike their system, answers in our case are expected to be an exact natural language response that aptly answers the question.

% In one of the early works on QA on videos, Yang et al.~\cite{old_news_videoqa}, introduced a system called VideoQA for answering factoid questions on videos. 
% Similar to our work, their dataset comprises news videos. However, unlike their dataset, 

% Yang et al.~\cite{old_news_videoqa} were the first to introduce a QA task for videos
% Similar to our work, their dataset comprises news videos and their system named VideoQA is designed to return  answers for simple factoid questions. In this early work, noisy OCR obtained from the videos was used to correct speech recognition errors. This correction resulted in better answer extraction. Contrary to the current work, scene text was not directly used for question answering.
%questions in~\cite{old_news_videoqa} did not require scene text understanding 

%The major difference between \cite{old_news_videoqa} and proposed work is that, questions in the proposed dataset require models to only read the textual content present in the video to obtain the answer.

More recent works in VideoQA %Majority of these works including 
\cite{moviesqa, tvqa, activityqa, refined_attention} require models to reason about the events taking place in videos, but disregard any textual information in the videos. Tapaswi et al.~\cite{moviesqa} introduced a dataset that aims to study story comprehension using video and subtitles. Zhou et al.~\cite{activityqa} introduced a large-scale VideoQA dataset that consists of videos of different activities. 
A method that gradually refines attention over the appearance and motion features is proposed in~\cite{refined_attention}, along with an automatically generated dataset for VideoQA using subtitles. 
%Authors in \cite{refined_attention} propose a method which gradually refines attention over the appearance and motion features which are guided by question. 
Yang et al.~\cite{just_ask} and Maharaj et al.~\cite{moviefib} focused on automatic generation of the VideoQA datasets. As the questions in \cite{just_ask} are automatically generated using captions, they are largely based on the visual appearance of objects and actions.  
% These works try to solve the obstacle of manual annotation by introducing methods which generate question-answer pairs automatically. 
Gupta et al.~\cite{newskvqa} explore knowledge-based question answering on news videos by proposing a new dataset. Questions in this dataset are primarily concerned with people seen in the videos, and the proposed models primarily rely on transcripts and an external knowledge base to find the answer. Questions in the above-mentioned works primarily require visual content and the transcripts of the videos to answer questions. Recently works such as \cite{hero, clipbert, blip, singularity} have introduced transformer-based models with different pretraining strategies and yield state-of-the-art performance on existing VideoQA datasets.
% are highly clustered around persons are require knowledge base and transcripts to obtain the answers.
%Recent works such as \cite{just_ask}, \cite{moviefib} have 

% have videos which require the models to reason over the visual events and commonsense reasoning occurring in the videos. 

%entities of person type. 
% One of the very early works over almost 20 years ago \cite{old_news_videoqa} highlighted the importance of exploring question-answer technique for news videos. The work included usage of multimodal features such as audio, video, textual information and external sources to enhance question-answering.

%Recently, authors in

% Questions in the above mentioned works, largely require visual content and the transcripts of the videos to answer questions. This becomes a motivation for the community to have publicly available video question answering dataset in which the questions require understanding of textual content in the video videos to obtain the answers.

Table \ref{tab:dataset_table} summarises existing works on VideoQA. It can be seen that majority of models focus on the visual content, transcripts and external knowledge to answer the questions. The text seen in the videos is an important source of information critical to understanding the content of news videos and videos shot  outdoors. However, existing works on VideoQA largely disregard text on the videos. This motivates the community to have a publicly available video question answering dataset in which the questions require understanding of the textual content in the videos to obtain the answers.

% On the contrary, questions in the proposed dataset require understanding textual content in the news videos to obtain the answers to the questions posed. 

\begin{figure}
    \centering
    \includegraphics[width=1.0\linewidth]{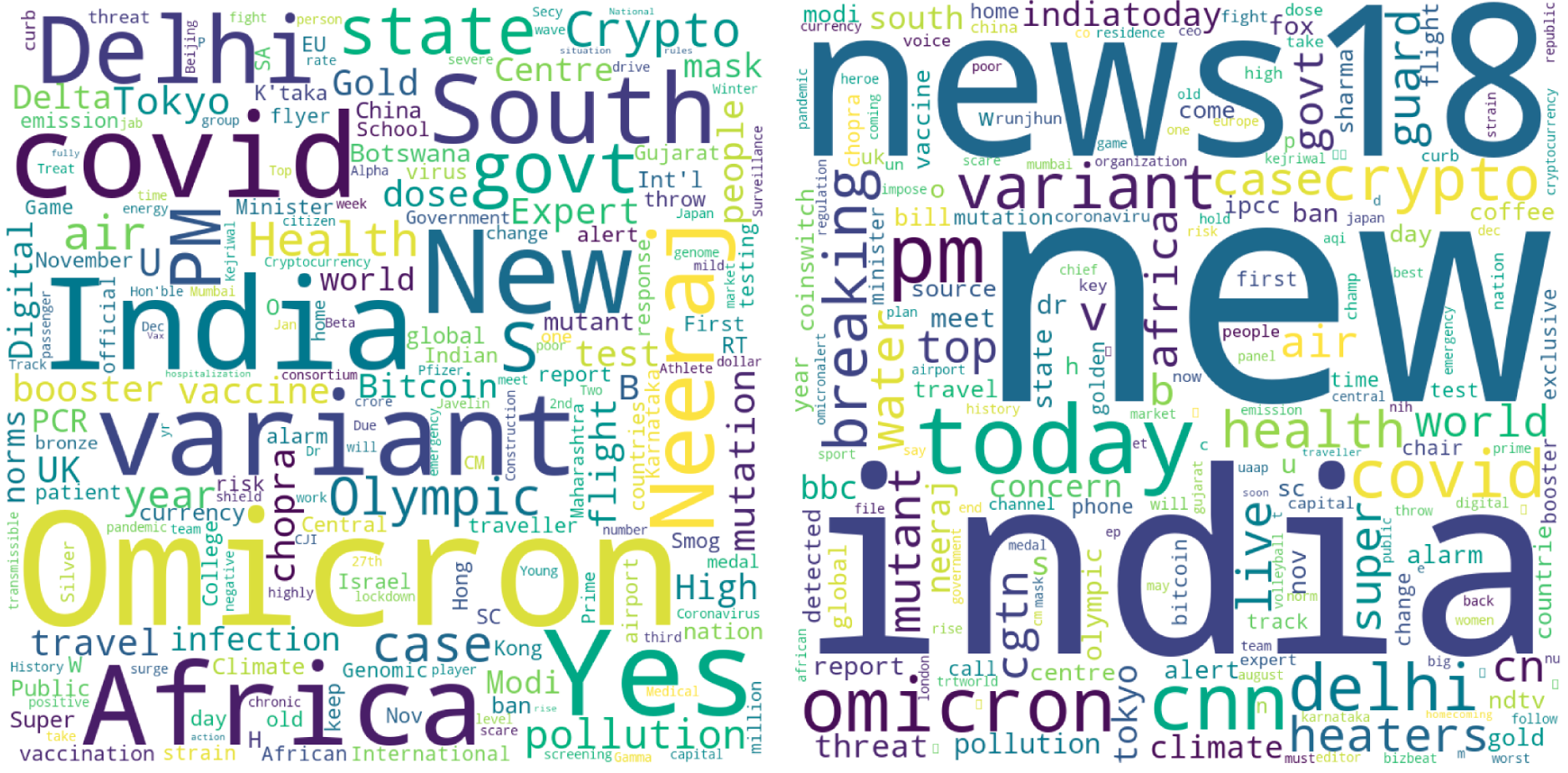}
    \caption{\small Word clouds of words in answers (left) and word clouds of words in OCR tokens (right).}
    \label{fig:word_clouds}
\end{figure}

\textbf{Scene Text Aware Visual Question Answering (VQA).} Early VQA  datasets for natural images mainly included questions that seek information present in the visual content of the images~\cite{vqa_survey}. However, realizing the importance of reading scene text to  understanding natural images, researchers have recently started working on VQA tasks for natural images where questions based on textual information in the images are prioritized. This VQA branch is referred to as Scene Text VQA. Two popular benchmarks for English scene text VQA are Scene Text VQA\cite{stvqa} and TextVQA~\cite{textvqa}. Wang et al.~\cite{vqa_evidence} extended scene text VQA to a bi-lingual setting by introducing a new dataset that contains images with English and Chinese scene text.
%For natural images, two popular benchmarks for VQA involving text are Scene Text VQA\cite{stvqa} and TextVQA~\cite{textvqa}. Wang et al.~\cite{}
% Realising the importance of  text for VQA---question answering on still images
% a number of VQA tasks that require reading text on the images, have recently been proposed. For VQA on natural images
% Scene Text VQA\cite{stvqa} and TextVQA~\cite{textvqa}
% recent works have been proposed 
% that incorporate the capacity to read image text in Visual Question Answering models.
%there have been many works lately that deal with Visual Question Answering (VQA) involving text.~\cite{stvqa,textvqa,ocrvqa,doccvqa}.
% These datasets contain questions asked on different types of images such as natural images, document images and im
% on real world images, book covers, document images. 
% Scene Text VQA\cite{stvqa} and TextVQA~\cite{textvqa} defined the reference datasets for this task.
%deal with VQA over natural images.
For scene text VQA, Singh et al.~\cite{textvqa} proposed a model called  LoRRA that uses top-down and bottom-up attention on scene text tokens and visual features to select an answer either from the OCR tokens or from a fixed vocabulary.
M4C~\cite{m4c} uses a multimodal transformer-based model for Scene text VQA and Text VQA. This model, unlike the LoRRA can generate answers of any length by combining tokens from a fixed vocabulary or the scene text tokens found on the image. The current state-of-the-art models for scene text VQA typically use a Transformer-based architecture that is trained in two stages; a pretraining stage and a finetuning stage~\cite{tap_2021,latr_biten_2022}. The pretraining stage in these works is designed to learn multimodal interactions.
In TAP~\cite{tap_2021}, Yang et al. propose to pretrain an M4C-like architecture using pretraining tasks suitable for alignment between scene text and visual objects. TAP uses visual features corresponding to visual objects detected on the images using a pretrained object detection model, as done in most of the previous VQA works like LoRRA and M4C.
Unlike TAP, which uses scene text, positional information, and visual features for pretraining on natural images, LaTr~\cite{latr_biten_2022} uses document images for pretraining and uses only text and layout information. In the finetuning stage, LaTr uses visual features extracted using a pre-trained vision transformer.

\begin{figure*}
\begin{subfigure}{.33\textwidth}
  \centering
  % include first image
  \includegraphics[width=.9\linewidth]{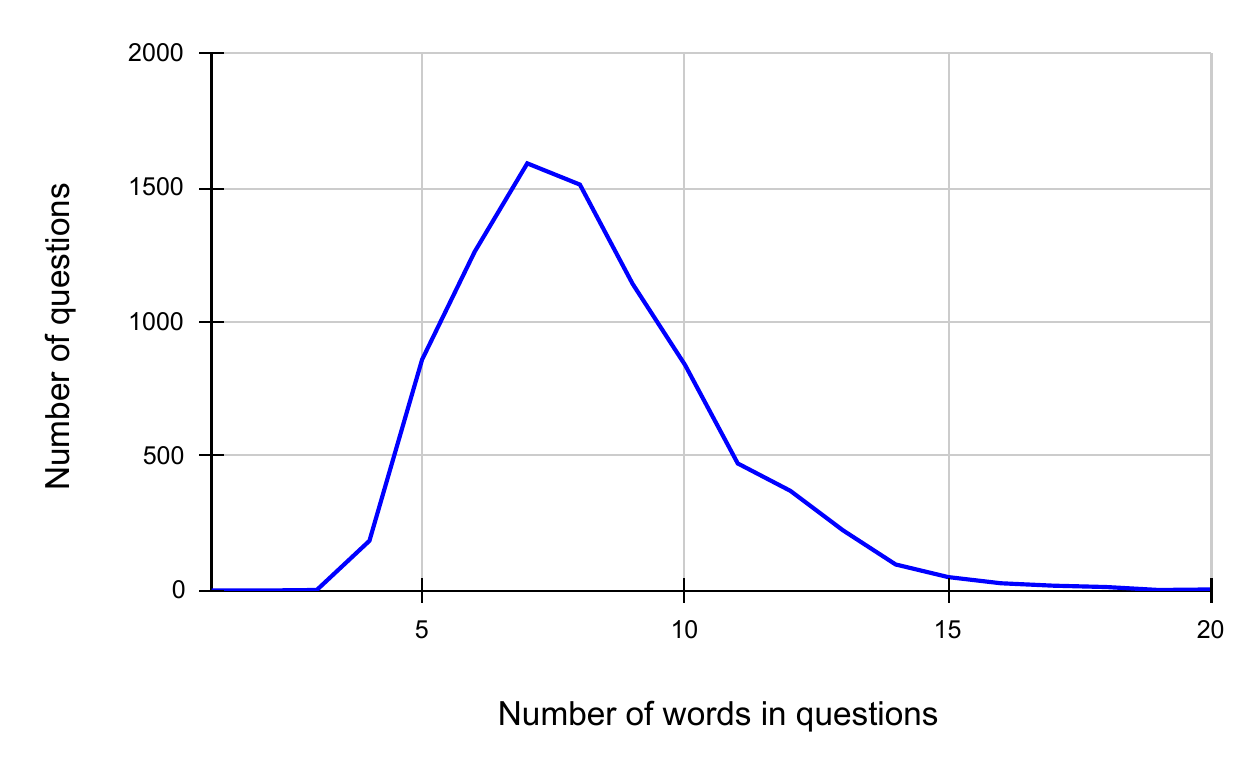}  
  \captionsetup{width=0.95\linewidth}
  \caption{\textbf{Questions with particular length.} The average length of questions in the dataset is $6.79$ words.}
  \label{fig:que_par}
\end{subfigure}
\begin{subfigure}{.33\textwidth}
  \centering
  % include first image
  \includegraphics[width=.9\linewidth]{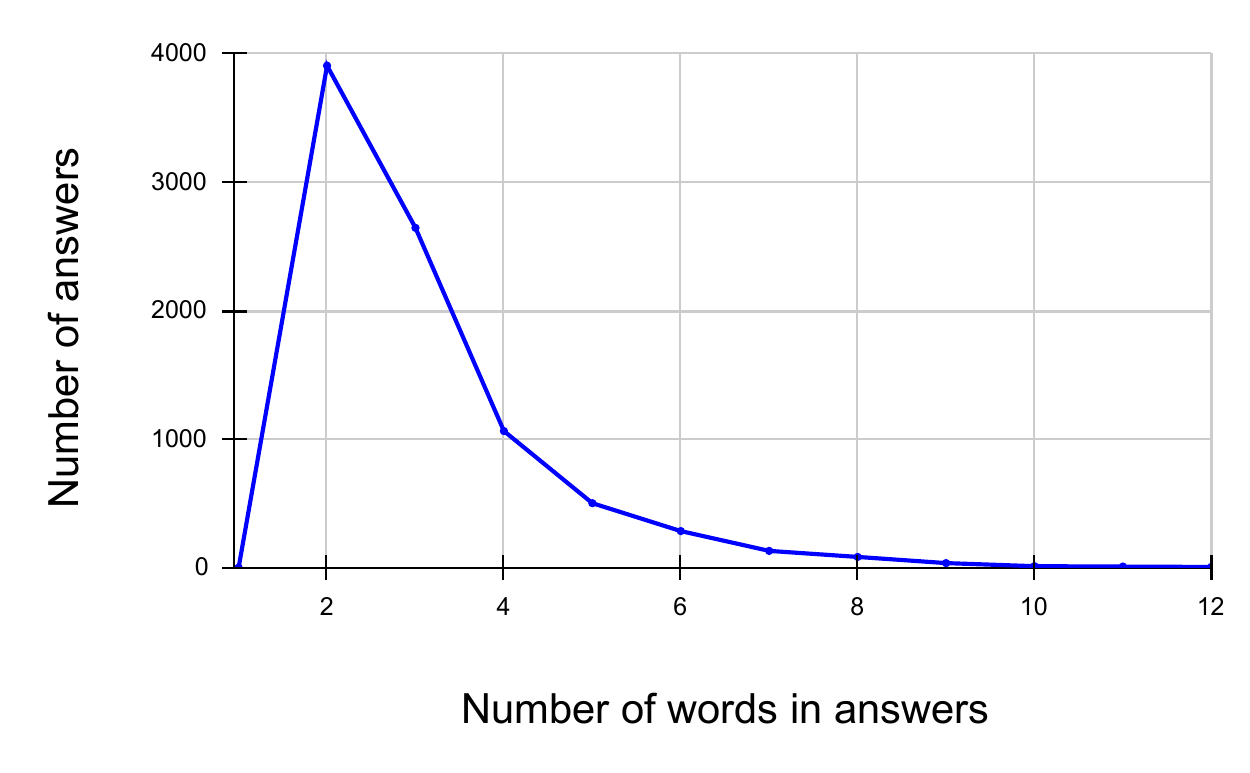}  
  \captionsetup{width=0.95\linewidth}
  \caption{\textbf{Answers with particular length.} The average number of words in the answers is $2.02$ words.}
  \label{fig:ans_par}
\end{subfigure}
\begin{subfigure}{.33\textwidth}
  \centering
  % include second image
  \includegraphics[width=.9\linewidth]{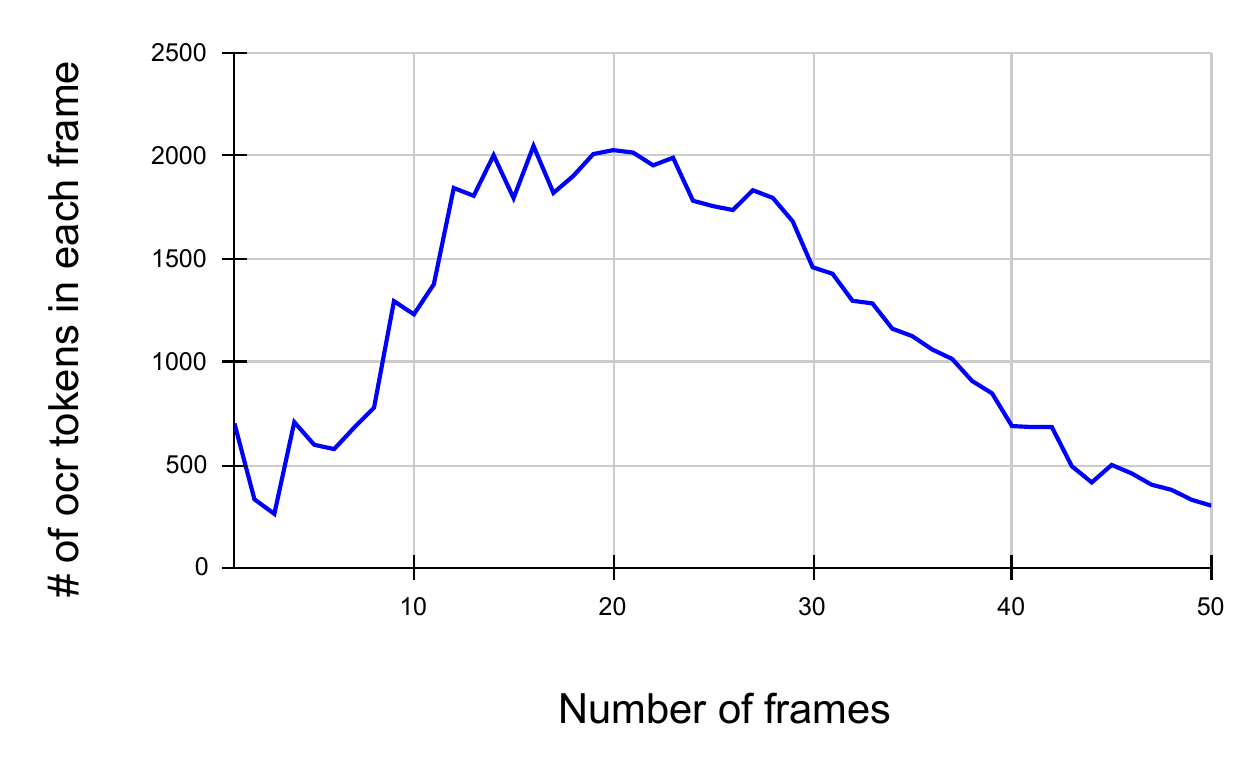}  
  \captionsetup{width=0.95\linewidth}
  \caption{\textbf{OCR tokens with particular length.} Average number of OCR tokens per frame is $26.14$ tokens.}
  \label{fig:ocr_par}
\end{subfigure}
\newline
\begin{subfigure}{.33\textwidth}
  \centering
  % include third image
  \includegraphics[width=.9\linewidth]{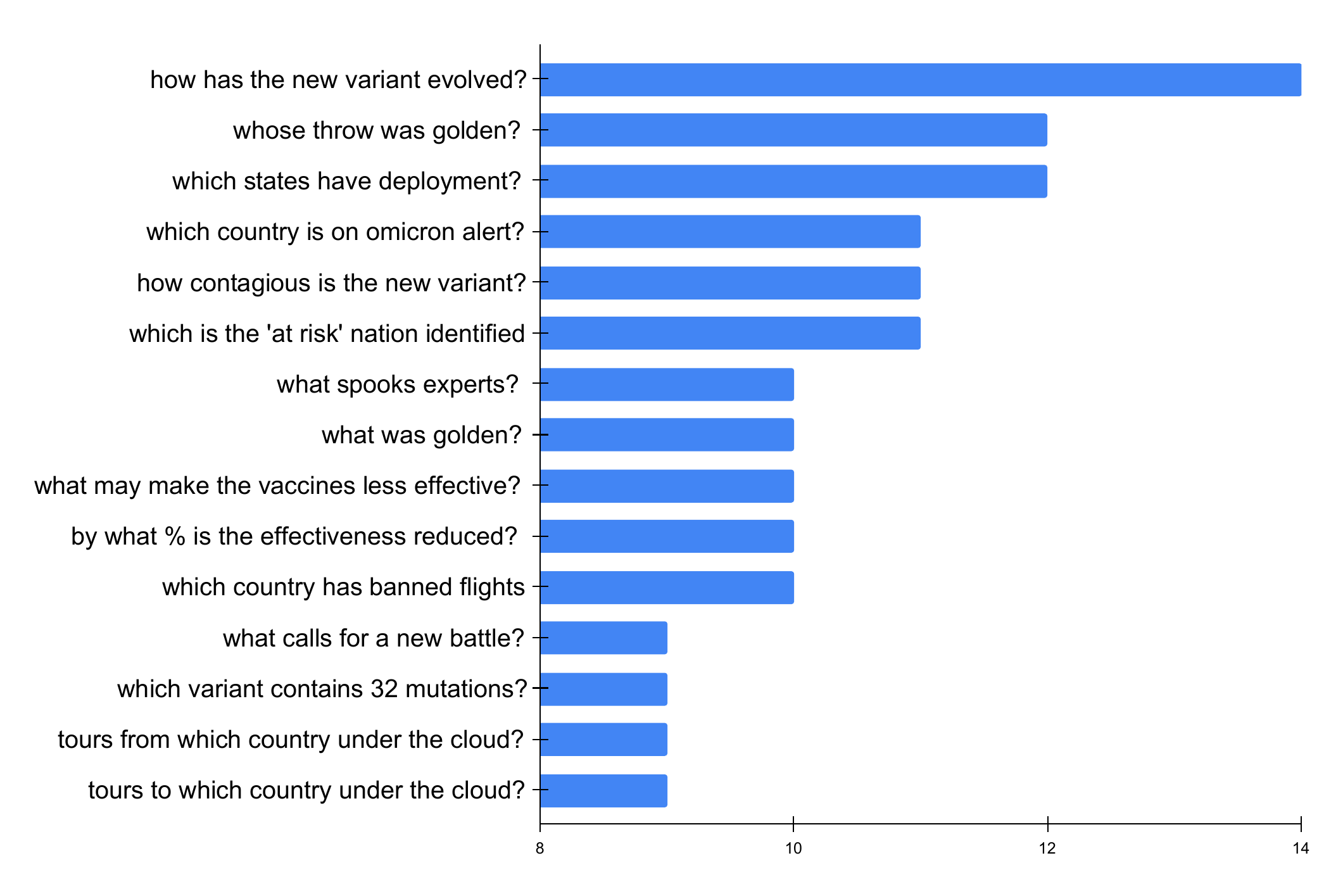}  
  \captionsetup{width=0.95\linewidth}
  \caption{Top 15 most occurring questions in the dataset.}
  \label{fig:top_que}
\end{subfigure}
\begin{subfigure}{.33\textwidth}
  \centering
  % include third image
  \includegraphics[width=.9\linewidth]{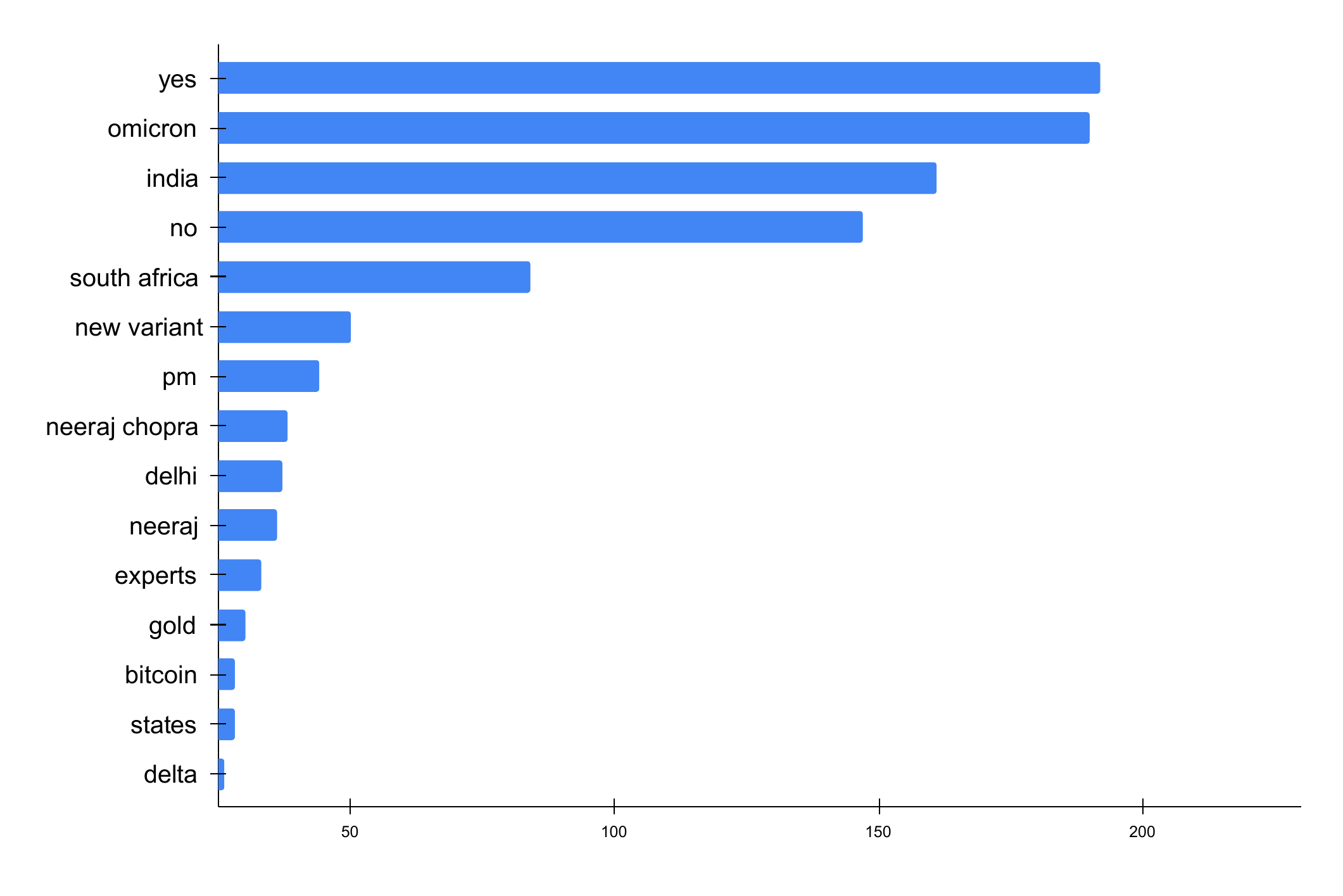}  
  \captionsetup{width=0.95\linewidth}
  \caption{Top 15 most occurring answers in the dataset.}
  \label{fig:top_ans}
\end{subfigure}
\begin{subfigure}{.33\textwidth}
  \centering
  % include fourth image
  \includegraphics[width=.9\linewidth]{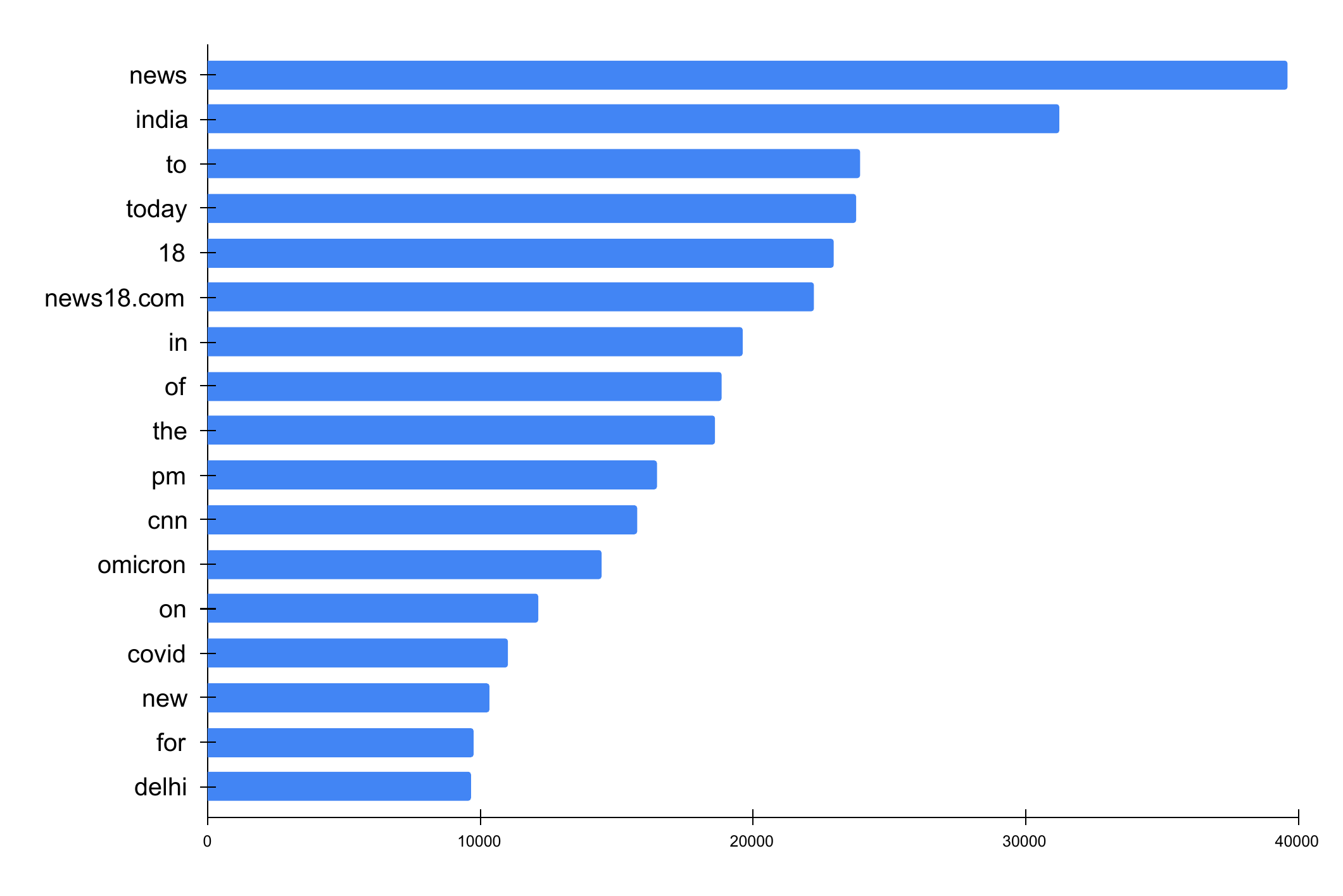}  
  \captionsetup{width=0.95\linewidth}
  \caption{Top 15 most occurring OCR tokens in the dataset.}
  \label{fig:top_ocr}
\end{subfigure}
\caption{Statistics for question, answer and OCR tokens in \textbf{NewsVideoQA dataset}.}
\label{fig:data_statistics}
\end{figure*}

% \begin{figure}
%     \centering
%     \includegraphics[width=1\linewidth]{images/question_bubble.pdf}
%     \caption{A bubble chart with number of words in question in horizontal axis and number of words in answers in vertical axis based on question type.}
%     \label{fig:que_bubble}
% \end{figure}

% Biten~\cite{latr_biten_2022} shows that for scene text VQA, pretraining on document images  using text and layout information alone (i.e. without using any visual features during pretraining) is 

% By combining the rich representation for text in images, this method fuses different modalities by projecting them into a common semantic space. 

In addition to scene text VQA, many specialized VQA tasks require reading and reasoning text on the images. There are multiple datasets for VQA on charts where text on the charts is critical to answer the questions~\cite{kafle2018dvqa,kahou2017figureqa}. Mishra et al.~\cite{mishra2019ocr} introduced a VQA dataset where all images are book covers, and the questions in the dataset are synthetically created using metadata associated with the books and question templates. Since the questions are created using  information such as author names, titles and names of the publisher, questions purely depend on the text on the book covers and need little visual information.   
DocVQA\cite{docvqa} extends VQA to text-rich document images. This dataset has questions grounded on various document elements such as unstructured text in the form of paragraphs, forms, tables, and figures.
% introduces a new dataset on question-answering for document images which includes multiple elements like tables, forms and figures. 

% Authors in \cite{doccvqa} introduce the problem of visual question answering over collection of documents by proposing a new dataset and task where the questions are imposed over the entire collection of document images. This setting as presented in \cite{doccvqa} where a collection of documents are used, is relatively similar to video question answering where multiple frames from a video are used for obtain answer.

Similar to these existing VQA works that involve text on still images, we propose a VQA task that requires reading and understanding the text on videos.
% VQA works that deal with VQA on still images containing embedded text, our work proposes a new Video QA task where text that is seen on the video is essential for finding the right answer for almost all of the questions.  
While the exact context---a single image---is given directly for the VQA problems, the proposed NewsVideoQA with multiple frames requires models to automatically find the right frames that informs the answer and reason over the textual content in those frames.

% These works presented in the literature restrict their scope of scaling to answer questions from videos using textual information present in the videos. In order to architect better multimodal comprehending systems, it is very essential for the methods to have the ability to read the text in the videos. Despite its primary importance, the task of text-based video question answering has not been tackled. We aim to fill this gap in the literature by introducing the novel task.

% \textbf{NEWSKVQA: Knowledge-Aware News Video Question Answering}

% Check also: 
% - "VideoQA: Question Answering on News Video" (very old paper, but worth mentioning that this was already attempted pre-deep, 20 years ago).
% - Do a subsection on generic VideoQA "TVQA: Localized, Compositional Video Question Answering" (https://arxiv.org/pdf/1809.01696.pdf) / "DramaQA: Character-Centered Video Story Understanding with Hierarchical QA" etc
% - Do a subsection on Scene Text VQA.

%--------------------------------------------------------------------------------------------------------------------------
%--------------------------------------------------------------------------------------------------------------------------

\section{NewsVideoQA Dataset}
\label{sec:dataset}
In this section, we explain the data collection and annotation process. Also, we share statistics and analysis of the proposed NewsVideoQA dataset.

\subsection{Data Collection}
\textbf{News Videos:} We collect news videos from English news channels around the world. We obtain videos from the following YouTube channels like BBC, ABC Australia, India Today, TRT World, AL Jazeera, CNN, NHK World Japan, Fox News, WION, NDTV, ABC News, CNN-News18, CTV News, CGTN, and IPCC. While collecting the news videos, we manually ensure that the videos are text-rich because the proposed task relies on video question answering, which requires reading text. The collected videos are split into 10 seconds of non-overlapping clips. The proposed dataset contains $3,083$ videos, with at least 20 videos from each channel. The average number of questions per video is $2.96$. The maximum number of questions defined for a video in the dataset is $20$. The minimum number of questions defined for a video is $1$.

\textbf{Questions and Answers:} The annotation process was organized into two stages. In stage 1, the annotators were instructed to define question-answer pairs based on textual information present in the news videos. Specifically, they were provided with the following instruction: \textit{`Ensure that answering the questions generated requires reading of the text present in the news videos and should be related to the topic of that video'}. Annotators were asked to frame factoid questions that can be answered by reading the text present in the news videos. They were also instructed to add a timestamp: the time (with up to 1 second precision) of the video when the question was framed.

A second stage of verification was introduced to check the correctness of the data.
% In addition, to correct any poorly annotated data, we introduce a verification stage. 
Here, the annotators were asked to verify the data collected in the first stage. The annotators were shown the video-question pair for a video clip, and were asked to enter the answer and the timestamp and check the correctness of the question-answer pair based on its relevance to the textual content of the news video. They were asked to reject the questions with any grammatical mistakes in the questions and answers. During this stage, if the annotator finds a question-answer pair irrelevant to the topic or if the question was framed on the audio of the news videos, then such question-answer pairs were rejected from the dataset. A total of $1,200$ QA pairs were rejected after the verification step. An extra stage was also added where the authors reviewed randomly picked question-answer pairs and their correctness and relevance to the task proposed.

\subsection{Statistics and Analysis}
The NewsVideoQA dataset comprises $8,672$ questions framed on $3,083$ news videos. The data is split randomly in 80-10-10 ratio to train, validation and test split. The train split has $6,994$ questions over $2,407$ videos, the validation split has $714$ questions over $330$ video clips, and the test split has $964$ questions over $346$ video clips. 

% Fig. \ref{fig:topic_dis} shows the topic-wise distribution of the question-answer pairs. It is observed that question-answers from Omicron virus (Corona virus) has the maximum percentage. This is due to the notable coverage in the news videos as we have collected these videos in specific time period where publicity of such topics in news videos was maximum.

\begin{figure}
    \centering
    \includegraphics[width=0.9\linewidth]{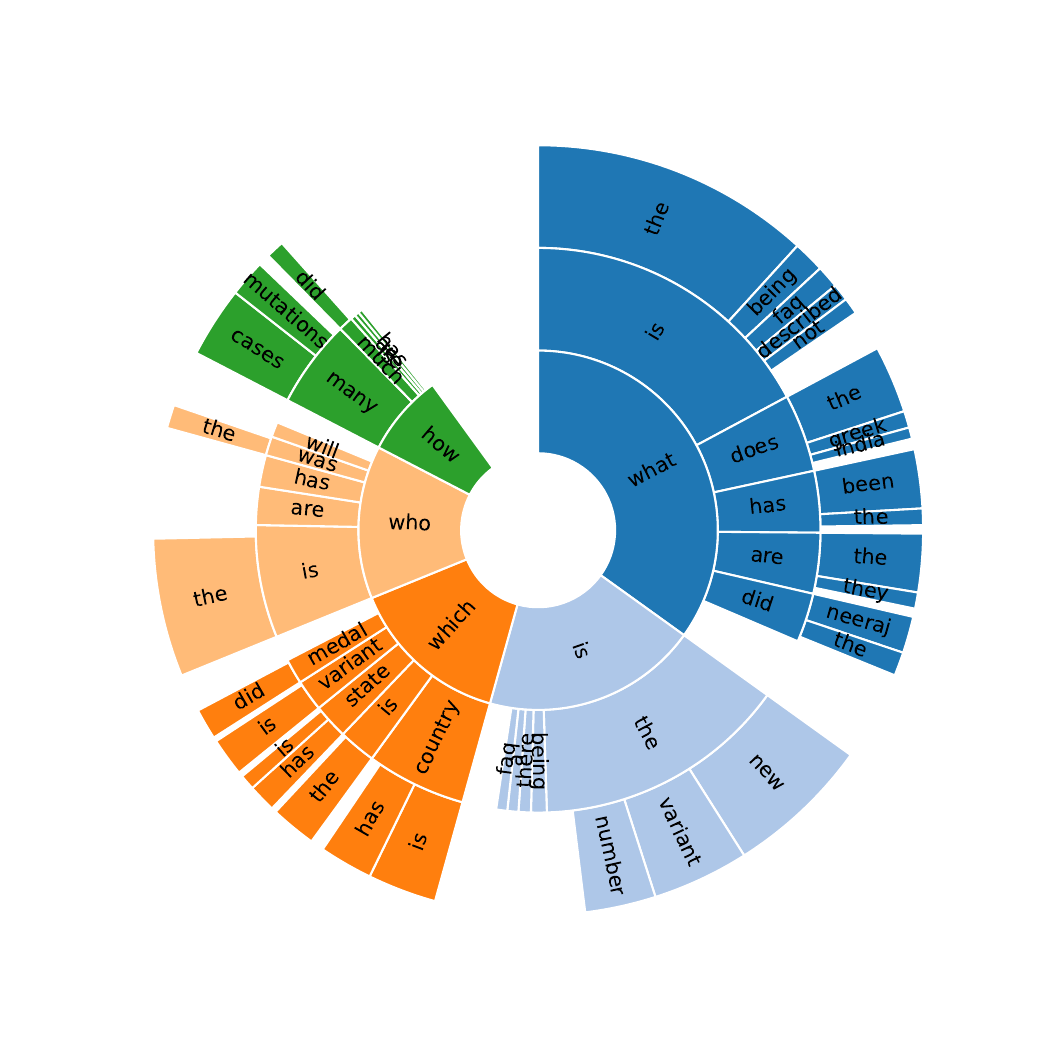}
    \caption{Distribution of questions by their starting 3-grams. Note that there is a diverse range of types of questions in the dataset. The question type \textit{"What"} has a maximum count with questions such as \textit{"What is the ...?"}, \textit{"What does the ...?"} and so on.}
    \label{fig:sunburst}
\end{figure}

Fig.~\ref{fig:que_par} shows the distribution of question lengths for the questions in NewsVideoQA dataset. The average question length is $7.04$ words. Among the $8,672$ questions $7,008$ ($80.81\%$) are unique. Higher diversity in questions is reflective of the fact that questions are based on textual content.
Fig.~\ref{fig:top_que} shows the top 15 most frequent questions and their frequencies. Fig.~ \ref{fig:sunburst} shows a sunburst plot of the first three words of the questions. It can be observed from Fig.~\ref{fig:sunburst} that there is variability in the question types like, questions starting with \textit{"What"} which account for the questions related to the answer being directly present in the text of news videos. We provide subtitles of the news videos  using a publicly available  speech-to-text tool\cite{silero}. A total of $1,388$ ($17.36\%$)  questions can be answered with sub-titles of the videos. This low percentage is observed due to two reasons, (a) smaller duration of the videos (10 seconds), resulting in incomplete sentences in the subtitles, and (b) most of the  questions are based on textual content of the news videos. In total, there are $4,150$ ($47.85\%$) unique answers. Word cloud on the right in Fig.~\ref{fig:word_clouds} shows the most common words in the answers. The answer space is broad and involves names of countries, events, games, people, etc. The distribution of answer lengths is shown in Fig. \ref{fig:ans_par}. The average answer length is $2.02$. The top 15 answers in the dataset are shown in Fig. \ref{fig:top_ans}. We obtain OCR tokens using Google OCR. We uniformly sample the video at 2 frames per second and also retain the first frame of the video. Fig.~\ref{fig:word_clouds} on the left shows the word cloud of OCR tokens. In Fig.~\ref{fig:top_ocr} we show the top 15 OCR tokens present in the dataset. An average of $26.14$ OCR tokens per frame is observed, and an average of $532.55$ OCR tokens per video clip are observed in the dataset. 

% The total number of questions that are framed based on the text on the videos is 7712 (96.26 \%). This suggests that reading text is important to answer the questions in our dataset. 
% In total, 186 (2.31 \%) questions require extra information from different parts of the video. In Fig.~\ref{fig:que_type_distribution} we present analysis of distribution of questions in NewsVideoQA dataset based on question types. 

\vspace{-0.2cm}

% We first convert the 3 minute video to 10 seconds clips on which the questions have been framed. We run a hashing based de-duplication algorithm \cite{dedup} to remove all the duplicate frames in the clipped video. Fig. \ref{fig:word_clouds} on the left shows the word cloud of OCR tokens. An average of 9 OCR tokens per frame in the video and an average of 160 OCR tokens per clip is present in the dataset.
% We analyze the OCR tokens produced by Google OCR. 

%--------------------------------------------------------------------------------------------------------------------------
%--------------------------------------------------------------------------------------------------------------------------

\section{Baseline Methods}
% In this section, we explain different baselines, where each originally designed baseline for a specific task is modified to incorporate the changes essential for the proposed task. We first evaluate several heuristic approaches, followed by machine comprehension models, Scene-text aware visual question answering models, and video question answering models.
%In this section we explain the baselines we use which include heuristic ones and trained models. We establish baselines performance metrics. 
We evaluate three different methods as strong baselines for the newly introduced task of scene-text aware VQA on NewsVideoQA dataset. In this section, we briefly discuss the original methods and explain how these methods are adapted for the new task.
% In this section, we explain different baselines that we use for the task of NewsVideoQA. We explain 
% and how these models that are originally designed for different but related tasks are adopted for 

%\subsection{Heuristics and Upper Bounds}
\subsection{Heuristic methods and Upper Bounds}
Inspired by heuristic baselines evaluated on scene text VQA~\cite{textvqa,stvqa} and DocVQA~\cite{docvqa} datasets, we evaluate the following heuristic baselines and upper bounds:
% \textbf{(i) Longest OCR token:} measures the performance when longest OCR token among all the frames is selected as the answer. 
\textbf{(i) Majority answer:} measures the performance when the most frequent answer in the train split is considered as the answer for all the questions in the test set. \textbf{(ii) Biggest OCR token:} measures the performance when the OCR token that occupies the largest area in the video is considered as the answer.
% (OCR token with maximum area, computed with the bounding-box of the OCR token). 

We compute upper-bound (UB)  for the following cases: \textbf{(i) Vocabulary UB:} measures the maximum performance obtainable on the test set, if an answer is picked from a vocabulary of most common answers in the train split. \textbf{(ii) OCR Substring of single frame UB:} this measures the performance that can be obtained when we restrict our vocabulary to list of OCR tokens of the frame on which the question was defined. \textbf{(iii) OCR Substring of all frames UB:} measures the performance we can obtain if the answer in the test split is a substring in the concatenated list of OCR tokens from uniformly sampled frames of the video.

% we can obtain for the questions in the test split, if the answer was present in the vocabulary of answers present in train split.measures the performance we can obtain if the answer in the test split is present as a substring in the concatenated the list of OCR tokens of a single frame in the video on which the question was framed.

% \textcolor{red}{TODOs,Biggest ocr token, longest ocr token from the same frame, ocr upper bounds, BERT finetuning on ocr list substring
% }

% \textbf{Random answer:} measures the performance when a random answer is chosen from the answers in the train split. \textbf{Random OCR token:} measures the performance when a random OCR token from the frames of a video is picked as the answer.

\subsection{Reading comprehension model}
\label{subsection:mrc}

As observed in~\autoref{sec:dataset}, by design, almost all of the questions in NewsVideoQA are grounded on the text in the videos. For this reason, we evaluate a QA baseline that only considers the text in the videos to answer the questions. Specifically, we evaluate the BERT \cite{bert} QA model that is originally developed for extractive text-only QA. Extractive QA is a task of extracting a short snippet from the document/context on which the question is asked. The answer snippet is called  a `span' and the span is defined in terms of its start and end tokens.
%In this case the answer is defined as a span in the context on which the question is defined. 
BERT is a transformer encoder-based method of pretraining language representations from unlabelled text. These pretrained models can be used later for downstream tasks with addition of output suitable for the task at hand. For the task of extractive QA, the additional layer, is an output layer that predicts start and end tokens of the span of the answer. For NewVideoQA, we concatenate the OCR tokens in a frame (assuming we know the correct frame) or the whole video---in our experiments we try out both settings---in the default reading order (i.e., top-left to bottom right order) and use this sequence as the context for the BERT QA model.

\subsection{VQA Model}
\label{subsection:vqa_models}

To evaluate the performance of current VQA models on NewsVideoQA dataset, we use M4C \cite{m4c} model which takes into account the text present in the frames of the news videos. We pair each question with the frame corresponding to the timestamp of the question defined and consider it as input to M4C. M4C uses a multimodal transformer and an iterative answer prediction module. The tokens in the questions are embedded using a BERT model \cite{bert}. Each frame is represented using the following features: (i) appearance features of the objects detected using a Faster-RCNN pretrained on Visual Genome \cite{visualgenome} and (ii) location information - bounding box coordinates of the detected objects. 

Each OCR token recognized from the frame is represented using the following features: (i) a pretrained word embedding, which is FastText \cite{fasttext}, (ii) appearance feature of the token's bounding box from Faster-RCNN \cite{fasterrcnn} (iii) PHOC \cite{phoc} representation of the token and (iv) bounding box coordinates of the token. The representations of the entities mentioned, i.e., question tokens, objects and OCR tokens are projected to a common, learned embedding space. Later, a stack of transformer \cite{attention_is_all_you_need} layers is applied over these features in the common embedding space. The multi-head self-attention in transformers enables both inter-entity and intra-entity attention. In the end, answers are predicted through iterative decoding in an auto-regressive manner. At each step in the decoding, the decoded word is either an OCR token from the considered frame or a word from the fixed vocabulary of the common answer words.

% Similar to the experiments defined in \ref{subsection:mrc}, we perform the similar experiments on M4C. (i) \textbf{Multiple frames + Voting.} We consider multiple frames with answer present in context, obtain the features essential for M4C, and perform voting based on answer occurring maximum number of times. (ii) \textbf{Single frame.} This case is similar to text-based VQA dataset where a single image along with its OCR tokens is considered as input to M4C.

\begin{figure}
    \centering
    \includegraphics[width=0.75\linewidth]{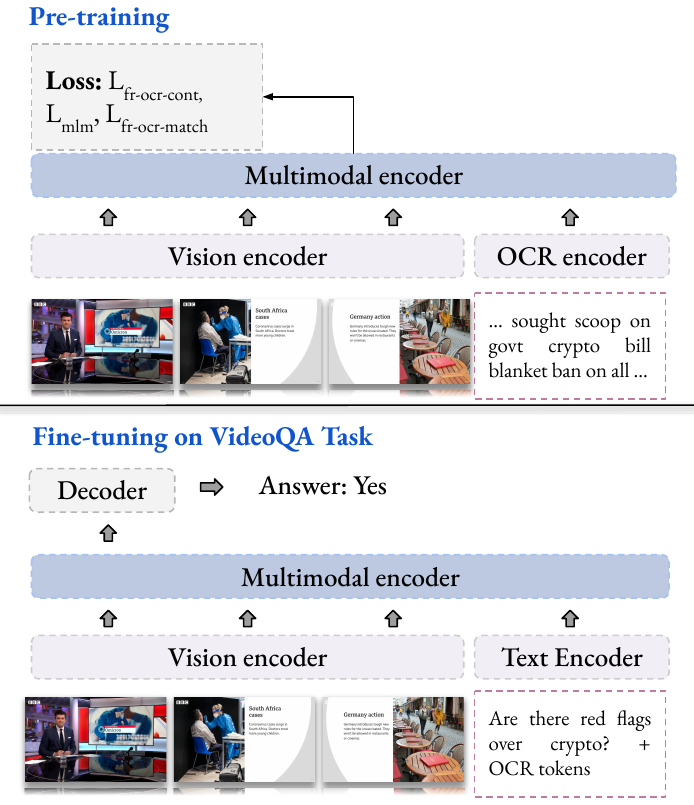}
    \caption{\small\textbf{OCR-aware SINGULARITY.} We extend SINGULARITY~\cite{singularity} for the task of text-based video question answering by incorporating OCR information by pretraining and finetuning on proposed NewsVideoQA dataset.}
    \label{fig:ocr_aware_videoQA}
\end{figure}

\subsection{VideoQA Model}

\setlength{\intextsep}{0cm}
\begin{table}
    \centering
    % \footnotesize
    \caption{\small\textbf{Heuristics and Upper bound baseline results.} It can be seen that answers are substrings for more than $50$ \% of the serialized OCR tokens of a single frame corresponding to the timestamp of the question.
    }
    \begin{tabular}{@{}lr@{}}\toprule
        \textbf{Heuristic Baselines} & \textbf{Acc. (\%)} \\
        \midrule
        Majority answer    & $3.00$  \\
        Biggest OCR token  & ${1.03}$  \\
        Vocab Upper Bound  & ${\mathbf{76.58}}$  \\
        Substring single frame UB  & $53.05$ \\
        Substring all frames UB & $74.43$  \\
        \bottomrule
    \end{tabular}
    \label{tab:heuristics_table}
\end{table}

In addition to the text-only QA model and the text-based VQA models, we evaluate  the performance of NewsVideoQA on a recently proposed transformer-based Retrieval and VideoQA method called SINGULARITY \cite{singularity}. This method studies the importance of temporal relations to answer questions. SINGULARITY is a vision-language model pretrained on many  video and image captioning datasets \cite{msrvtt, didemo, activity_net_captions, refined_attention, activityqa, msrvtt_mc}. It consists of three components, a vision encoder \cite{vit}, a language encoder \cite{bert} and a multi-modal encoder \cite{attention_is_all_you_need}. For pretraining, each video/image is paired with its corresponding caption. The multi-modal encoder applies cross-attention to collect information from visual representations using the text as the key. Three pretraining objectives are defined: (i) Vision-Text Contrastive: a contrastive loss that aligns vision and text representations, (ii) Masked Language Modeling (MLM): predicts the masked visual and text contexts, and (iii) Vision-Text Matching: predicts the matching score of a vision-text pair with multi-modal encoder. For QA task, a multi-modal decoder is initialized from pretrained multi-modal encoder, which takes the outputs of multi-modal encoder as input. This generates an answer text with "[CLS]" as start token.

% As the proposed dataset is a video question-answering dataset, we evaluate the performance of recently proposed transformer based VideoQA method SINGULARITY \cite{singularity} on NewsVideoQA dataset. 

% A standard vision-language model, with vision encoder for frame and text encoder for text encoding are used to represent the input. These encoders are followed by multi-modal encoder. This multi-modal encoder is trained on cross-attention for cross-modal fusion. It follows pre-training on large datasets followed by sparse frame model for VideoQA.
% As, this work proposes standard video question answering method using transformers, we evaluate the performance of this model on the proposed dataset. To extend this model to include OCR tokens we pair the news videos with OCR tokens of the frames on which a question is framed.

We extend the original SINGULARITY model \cite{singularity}, Fig.~\ref{fig:ocr_aware_videoQA} and propose a new \textbf{OCR-aware VideoQA} version that can read the text in the videos and thereby answer questions based on the text in the videos. 
To this end, we include the OCR tokens in the videos as additional input during  pretraining and finetuning stages. At the time of pretraining, unlike the original model that uses   image/video + caption pairs, we use  image/video + OCR tokens pairs. Similar to the original model, the following three pretraining objectives are employed.

\begin{table*}
    \centering
    \caption{\textbf{Comparison of all baselines on test set:} It can be seen that models such as BERT-QA \cite{bert} have poor performance when input of 12 frames followed by voting is provided at the test time. SINGULARITY \cite{singularity} without any OCR information performs very poor as it does not consider OCR tokens as input. OCR-aware SINGULARITY performs better than all the baselines.} 
    % \begin{tabular}{@{}p{2.5cm}p{0.5cm}p{1cm}p{1.3cm}p{1cm}@{}}\toprule
    % \resizebox{\columnwidth}{!}{%
    \begin{tabular}{@{}lccrr@{}}\toprule
        \textbf{Model} & \textbf{\#Frames for training} & \textbf{\#Frames for testing} & \textbf{{Acc. (\%)}} & \textbf{ANLS} \\
        \midrule
        BERT-QA \cite{bert} & 1 & 1 (1 frame from the video) &${28.70}$ & ${34.21}$\\
        BERT-QA \cite{bert} & 1 & 1 (1 frame on which question was defined) &${46.55}$ & ${56.81}$\\
        M4C \cite{m4c}  & 1 & 1 & $28.49$ & $32.17$ \\
        BERT-QA \cite{bert} & 1 & 2 (2 frames from the video) &${15.03}$ & ${17.65}$\\
        BERT-QA \cite{bert} & 1 & 2 (2 frames on which question was defined)  &${56.36}$ & ${67.11}$\\
        M4C \cite{m4c}  & 1 & 2 & $27.87$ & $31.54$ \\ \midrule
        % BERT-QA \cite{bert} & 1 & 12 & ${19.61}$ & ${25.88}$\\
        BERT-QA \cite{bert} & 1 & 12 (OCR tokens from 12 random frames) & ${53.86}$ & ${65.27}$\\
        M4C \cite{m4c} & 1 & 12 & $30.68$ & $34.90$ \\
        SINGULARITY \cite{singularity} & 1 & 12 & $4.82$ & $5.78$  \\
        OCR-aware SINGULARITY & 1 & 12 (OCR tokens from a single frame) & ${{33.57}}$ & ${{37.52}}$ \\
        OCR-aware SINGULARITY & 1 & 12 (OCR tokens from 12 random frames) & ${{32.47}}$ & ${{35.56}}$ \\
        \bottomrule
    \end{tabular}
    % %
    % }
    \label{tab:supp_ablation_table_all_baselines}
\end{table*}

\textbf{(i) Vision-OCR Contrastive loss:} aligns the visual features and OCR tokens,
%embeddings by maximizing the alignment between corresponding multi-modal embeddings,
\textbf{(ii) Masked Language Modeling:} follows the formulation in BERT\cite{bert} to predict a randomly masked OCR token and \textbf{(iii) Vision-OCR Matching:} similar to Vision-OCR Contrastive loss, this allows the models to improve the alignment between paired vision and OCR inputs by using output of [CLS] token from multimodal encoder for binary classification. In essence, it says whether or not the input frame and OCR tokens pair match. Similar to the original model, we add multimodal decoder that has same architecture as that of multimodal encoder. This decoder uses multimodal encoder outputs as its cross-attention inputs. It decodes the answer with [CLS] as the start token.

% At the time of finetuning, we provide OCR tokens as extra input and the rest is similar to how the original model is fine-tuned for video VQA.

% to improve the alignment between the paired vision and generate answers based on using decoder as defined in \cite{singularity}.

% It gradually refines the attention over the appearance and motion features of the video by considering the question as a guiding factor. A question is proposed word by word until the model generates the final optimized attention of the video features over question tokens. The weights obtained after refinement are used to generate the answer. The models is able to refine both coarse-grained question feature and fine-grained word feature together as guidance. The method contains Attention Memory Units (AMU) that transforms and manipulates the attention by processing each word and refines the attention over appearance and motion features of the video at each timestep.

\section{Experiments}
In this section, we explain evaluation metrics, and experimental settings and report the results. In all the experiments, we use the validation split of the dataset to save the best-performing checkpoints.

\subsection{Evaluation Metrics}
We use two evaluation metrics---Accuracy (Acc.) and Average Normalized Levenshtein Similarity (ANLS).  
Accuracy  is the percentage of questions for which the predicted answer matches exactly with the target answer. The accuracy metric awards a zero score even when the prediction is a little different from the target answer.
ANLS is a Levenshtein Similarity-based metric that acts softly on minor answer mismatches that might stem from an error in recognizing text on the images (i.e., OCR errors). Since all  the answers in our dataset are derived from text seen in the videos, we found ANLS to be a suitable metric for NewsVideoQA.

% ANLS act softly on minor asnwer mismatches 
% we show the that was proposed for evaluation on the ST-VQA dataset \cite{stvqa}. We use ANLS as it does not severely penalise small answer mismatches occurring due to OCR errors.
% minor answer mismatches stemming from
% OCR errors are not severely penalized neglects slight answer mismatches caused by OCR errors, and hence the predictions are not significantly penalised.

\begin{figure*}
    \centering
    \includegraphics[width=0.8\linewidth]{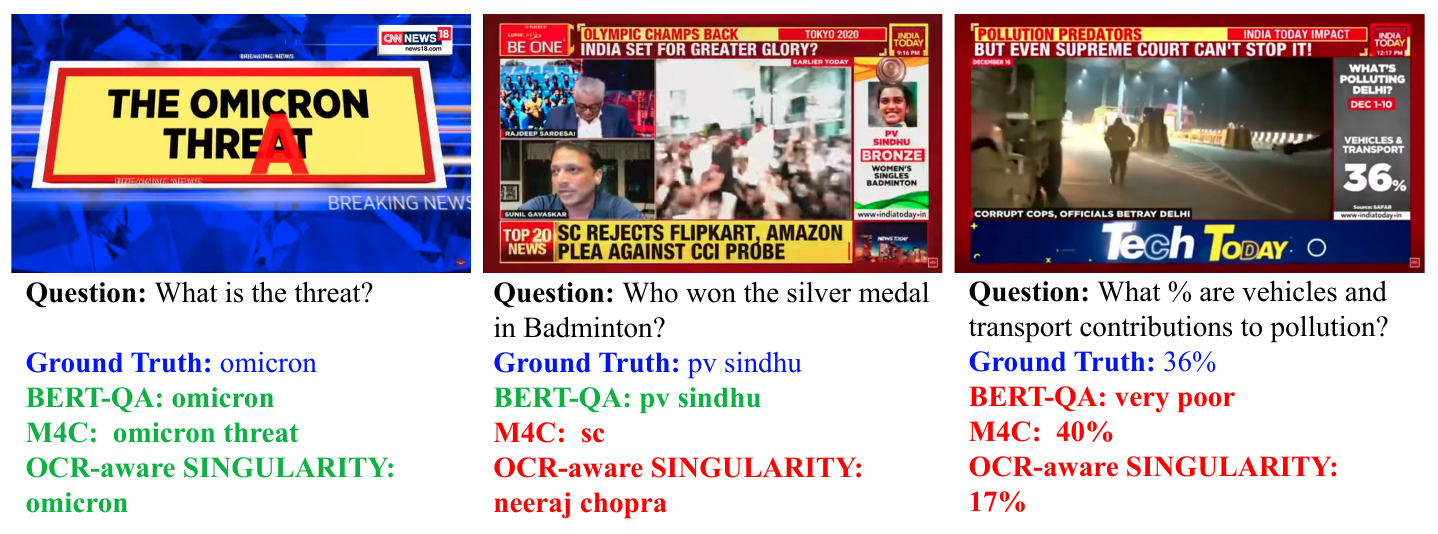}
    \caption{\textbf{Qualitative results} for different baselines on the proposed task. Results for baselines are shown in green for the correct predictions and in red for the incorrect predictions.}
    \label{fig:qualitative_results}
\end{figure*}

\subsection{Experimental setup}
\label{subsection:experimental_setup}
We run a commercial OCR engine to obtain OCR tokens for the evenly sampled frames. 

\textbf{BERT-QA.} In the case of NewsVideoQA, we use the OCR tokens of the sampled video frames as context for BERT-QA. For each question, we obtain the OCR tokens of the frame on which the question is defined. We use the default OCR token ordering from the OCR system: top-left to bottom-right. To convert the NewsVideoQA dataset in SQuAD format, we find the first substring of the answer in the context, which is an approximation of the answer span as followed in \cite{docvqa}. We finetune the BERT QA checkpoint that is already pretrained and finetuned for QA on SQuAD dataset~\cite{squad}. Specifically, we use the `bert-large-uncased-whole-word-masking-finetuned-squad' checkpoint~\cite{huggingface}.
%the pretrained BERT-QA SQuAD, (bert-large-uncased-whole-word-masking-finetuned-squad) 
We train the BERT QA model starting from this checkpoint  on NewsVideoQA dataset for ten epochs with a batch size of $32$ and a learning rate of $2e-05$. 
% We perform several ablation studies on BERT-QA, which are discussed in \ref{subsubsection:ablation_study}. 

\textbf{M4C.} For M4C, we use the official implementation along with default hyperparameters~\cite{mmf}.
The fixed vocabulary used for answer generation is  $3,751$  words from answers in the train split of NewsVideoQA. Since M4C is a model for VQA on images, we train it using  video frame + question pairs  in the train split of NewsVideoQA. Similar to how BERT-QA was trained on NewsVideoQA, for each question the corresponding matching frame is found using the time-stamp information for each question that was collected during annotation.
%Similar to BERT-QA,  for each question, based on the question time-stamp we consider the frame of the video based on the timestamp of the question as an input to the model. 
% We perform several ablations on M4C, which are discussed in \ref{subsubsection:ablation_study}. 

\textbf{SINGULARITY\cite{singularity}.} We use the pretrained model of SINGULARITY and finetune it on NewsVideoQA. We finetune it for 20 epochs and all the hyperparameters and training settings are kept the same as in the official implementation. SINGULARITY uses a single frame while training, and 12 randomly sampled frames while testing.

\textbf{OCR-aware SINGULARITY.} We continue pretraining the original  SINGULARITY on our NewsVideoQA dataset for 10 epochs. The vision encoder and the multimodal encoder are initialized similar to the original work.
% As in original singularity,  we initialise vision encoder using BEiT\textsubscript{BASE} \cite{beit} model trained on ImageNet-21K \cite{image_net_21k}. The multi-modal encoder is also initialised similar to how it is done in the original work. 
Following the pretraining on NewsVideoQA, 
we finetune the pre-trained model for $20$ epochs with a batch size of 4 and learning rate of $1e-5$. The only difference compared to the original model is that we append OCR tokens to the question tokens.
We keep the hyperparameters and pretraining settings the same as the SINGULARITY \cite{singularity}. More details on experimental settings for above mentioned baselines can be found in supplementary material. In order to maintain the constant setting throughout all the baselines, (multiframe at the time of testing), for BERT-QA and M4C we perform additional experiments where these models are trained on single frame and are tested on multiple randomly sampled frames followed by a majority answer voting to obtain the final answer. Similar to SINGULARITY, we fix the number of frames used at the time of testing to be equal to 12.

\subsection{Results}
\label{subsection:results}

\label{subsubsection:quant_res}

In Table.~\ref{tab:heuristics_table}, we show the results of heuristics and upper-bound baselines.  $3.0\%$ of the questions can be answered by predicting ``yes" which is the most common answer in the train split. Vocab Upper Bound of $76.58\%$  shows that many answers in the train split repeat in the test split as well. In Table.~\ref{tab:supp_ablation_table_all_baselines}, we show the comparative results for all the baselines. From the first four rows in the Tab.~\ref{tab:supp_ablation_table_all_baselines}, it can be seen that BERT-QA (text only model), and M4C (Text-based single image VQA model) perform good when they are tested on one frame and two frames settings (frames based on the timestamp of the question). It can be seen that the performance of M4C reduces significantly when it is tested on 12 frames. For BERT-QA, the first row shows the performance when OCR tokens a single frame (not necessarily the frame on which the question was defined). This is followed by testing BERT-QA on OCR tokens of the frame on which the question was defined. From the table, it can be seen that M4C performs poorly when the correct information required to answer the questions is not given as input to these models. SINGULARITY (without finetuning on NewsVideoQA) has poor performance compared to other baselines as the majority of the questions framed are based on textual content in the videos (Note: Results for BERT-QA in the first version were on OCR tokens not from the frames on which question was defined which led to poor performance of BERT-QA). We perform several experiments on the baselines which are present in the supplementary material. In Fig.~\ref{fig:qualitative_results}, we show qualitative results from our experiments. The left example shows the predictions of baselines. As the frame contains less textual information all the baselines predict the correct answer. Whereas in the center and right example, the number of OCR instances increases thereby increasing the difficulty to obtain the correct answer.

\section{Conclusion}

We introduce and explore the problem of text based Video Question Answering, in which the models are encouraged to read and reason about the textual content in the videos. Towards this, we propose a new dataset, NewsVideoQA, which contains questions defined over textual content in news videos. We adopt existing baselines for text based video question answering on NewsVideoQA. Furthermore, we redesign the existing VideoQA method by incorporating OCR tokens to yield better results compared to the original method. Our exhaustive analysis and findings encourage the concurrent use of visual and textual cues for better video understanding systems. Our work will encourage researchers to develop better text-based Video Question Answering models and better insights into well-designed multimodal machine understanding models.

% We introduced the novel task of `Text-based Video Question Answering' where given a news video, the task is to obtain an appropriate answer by reading text present in the video. We introduce a new dataset called NewsVideoQA for this task. We also provide results for baseline methods that showcase the performance of existing methodologies. We propose the first OCR-aware VideoQA model that can be the potential starting step in the direction of VideoQA models that can read. This contemporary task will encourage researchers to develop better text-based Video Question Answering models and better insights into well designed multimodal machine understanding models.provide detailed experiments on the baselines to showcase the performance of existing methodologies on NewsVideoQA. 

\textbf{Acknowledgements}
This work is supported by MeitY, Government of India.

{\small
\bibliographystyle{ieee_fullname}
\bibliography{egbib}
}

\end{document}